\pdfoutput=1

\documentclass[11pt]{article}

\usepackage[]{acl}

\usepackage{times}
\usepackage{latexsym}
\usepackage[size=tiny]{todonotes}
\usepackage{enumerate}
\usepackage[shortlabels]{enumitem}
\usepackage{amsmath}
\usepackage{booktabs}
\usepackage{graphicx}
\usepackage[normalem]{ulem}
\useunder{\uline}{\ul}{}
\usepackage{multirow}
\usepackage{subfigure}
\usepackage{float}
\usepackage{hyperref}
\usepackage{stfloats}
\usepackage{bbding}
\usepackage{tabularx}
\usepackage{colortbl} 
\usepackage{xcolor}
\usepackage{textpos}

\newcommand{\funny}[0]{\emph{FUNNY}}
\newcommand{\nfunny}[0]{$\neg$\emph{FUNNY}}
\newcommand{\medfunny}[0]{\emph{FUNNY}$_{\text{med}}$}

\newcommand{\RN}[1]{%
  \textup{\uppercase\expandafter{\romannumeral#1}}%
}

\newcommand{\yc}[1]{\textcolor{black}{#1}}

\newcommand{\se}[1]{\textcolor{black}{#1}}
\newcommand{\sen}[1]{\textcolor{black}{#1}}
\newcommand{\ycn}[1]{\textcolor{black}{#1}}
\newcommand{\serev}[1]{\textcolor{black}{#1}}
\newcommand{\ycrev}[1]{\textcolor{black}{#1}}
\newcommand{\rmd}[1]{\textcolor{black}{#1}}

\definecolor{y}{rgb}{1,1,0.4} 

\usepackage[T1]{fontenc}

\usepackage[utf8]{inputenc}

\usepackage{microtype}

\title{Transformers Go for the LOLs: \\
Generating (Humourous) Titles from Scientific Abstracts End-to-End}

\author{
  Yanran Chen, Steffen Eger \\
  Natural Language Learning Group (NLLG) \\
  University of Mannheim, Germany\\
  \texttt{yanran.chen@stud.tu-darmstadt.de} \\ \texttt{steffen.eger@uni-mannheim.de} \\
  }

\begin{document}
\maketitle
\begin{abstract}
  We consider the end-to-end abstract-to-title generation problem, exploring seven recent transformer based models (including ChatGPT)  fine-tuned on more than 30k abstract-title pairs from NLP and machine learning (ML)  venues. As an extension, we also consider the harder problem of generating humorous paper titles. For the latter, we compile the first large-scale humor annotated dataset for scientific papers in the NLP/ML domains, comprising 
  \yc{$\sim$2.6k} titles. We evaluate all models using human and automatic metrics.  
  Our human evaluation suggests that our best end-to-end system performs similarly to human authors (but arguably slightly worse). Generating funny titles is more difficult, however, and our automatic systems clearly underperform relative to  
  humans and often learn dataset artefacts of humor. 
  Finally, ChatGPT, without any fine-tuning, performs on the level of our best fine-tuned system.\footnote{Our paper title is a (modified) merge of a funny and unfunny title suggested by ChatGPT (\url{chat.openai.com}). Our paper logo is drawn by DALL-E (\url{https://openai.com/dall-e-2/}).\\Data+code: \url{https://github.com/cyr19/A2T}}
\end{abstract}

\begin{textblock*}{1cm}(12.5cm,-13cm) 
\includegraphics[scale=0.08]{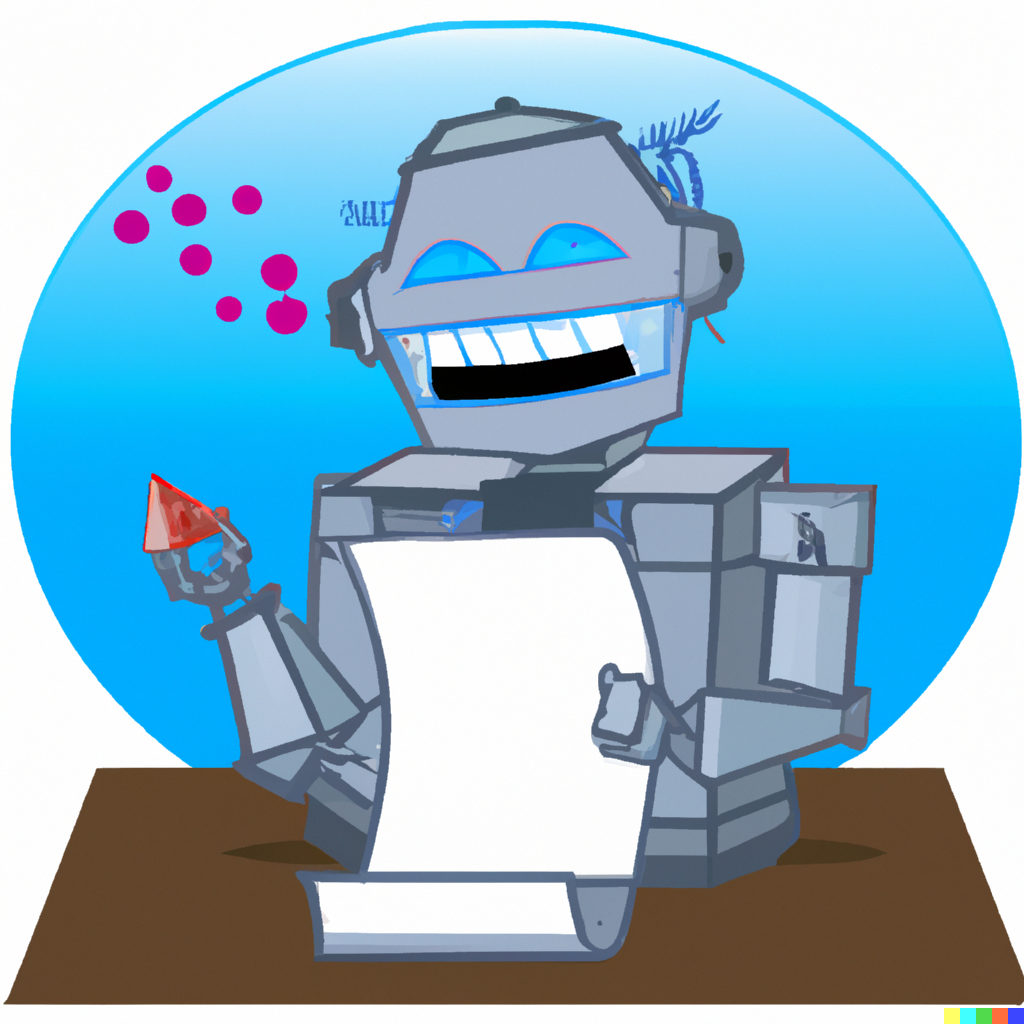}

\end{textblock*}

\section{Introduction}\label{sec:intro}

Computer-assisted writing is an important and long-standing use case of 
NLP 
and natural language generation (NLG) \citep{Burns1979StimulatingRI}, \sen{e.g., via and beyond tools such as spell checkers or grammatical error correction.}
The recent success of large-scale language models (LLMs), such as the GPT generation of NLG models, 
has made the goal even more   
realistic and promises full-scale automatic text generation, without any human intervention. 

In this work, we concern ourselves with 
automatic text generation in the scientific domain. 
Sample scenarios in this general context involve (semi-)automatically generating reviews for scientific papers \citep{yuan2022can}, e.g., as a response to high reviewing load in the face of exploding submission numbers; and generating captions for tables that require reasoning capabilities \citep{moosavi2021scigen}. Our goal is much more modest: we ask whether language models can generate adequate titles given a human authored abstract as input; we refer to this task as \textbf{A2T} (abstract-to-title generation). Title generation is important as titles are the first access points to  
papers; a good title may  
attract more readers and consequently increase paper 
impact, e.g., in terms of citation numbers \citep{falagas2013impact}. Besides generating titles per-se, we also aim for generating \emph{humorous} titles, an inherently difficult problem due to small sample size \sen{and the vagueness of humor}. 
Generating funny titles may be relevant as a funny title may attract even more readers: 
\sen{for example,  \citet{Heard2022IfTT} find that funny titles 
have significantly higher citation rates.}

We approach the problem as a standard sequence-to-sequence text generation problem, where we fine-tune 
LLMs 
on more than 30k abstract-title pairs from 
ML and NLP. 
Our contributions:
\begin{itemize}[topsep=5pt,itemsep=0pt,leftmargin=*]
  \item 
  \textbf{(i)}
  We provide the first publicly available humor annotated dataset for scientific titles in the NLP and ML domain,  
  with 
  \yc{2,638}
  humor annotated titles annotated by 2 annotators with decent levels of agreement (kappa $\sim$0.65). 
  \item 
\textbf{(ii)}
  We explore 6 recent popular text generation systems on the A2T task, finding one 
  to be competitive to human titles, according to automatic and human evaluation involving 15 annotators.
  \item 
\textbf{(iii)}
We analyze the problem and find that the A2T task is to some degree ill-posed as a good title may leverage more than the abstract alone (we argue that the problem framing is still a legitimate and efficient approximation). 
  \item 
\textbf{(iv)}
  For humor generation, 
  we find that our models clearly underperform relative to humans and instead often learn dataset artefacts.  
  \item 
  \textbf{(v)}
  We finally analyze ChatGPT on a small scale and find that it may be competitive to (albeit slightly weaker than) our best fine-tuned model without any task-specific fine-tuning at all. 
\end{itemize}
\section{Related Work}\label{sec:related}

\paragraph{Title generation and evaluation}
\citet{mishra2021automatic} perform A2T with pre-trained GPT-2 fine-tuned on arxiv papers and subsequent (rule-based) modules of title selection and refinement. 
We compare many more text generation models for the task, use better evaluation 
(including more comprehensive human and automatic evaluation), do not make use of rule-based selection and also consider humor in title generation. 
\citet{putra2017automatic} classify sentences from paper abstracts into rhetorical categories, retain those relating to methods and results and then generate titles using templates. They further note the relationship between the task of summarization \citep{nenkova2011automatic} and A2T, as a title can be seen as a summary of the research paper. We also leverage the relationship to summarization by considering pre-trained models fine-tuned on summarization datasets. In contrast to \citet{putra2017automatic} and \citet{mishra2021automatic}, we only consider end-to-end models that do not involve  pipelines. \sen{While refinement steps could be further helpful (but also error-prone), they additionally require  potentially undesirable  
human intervention \citep{Belouadi2022ByGPT5ES}.} \se{Related to the task of title generation is the task of headline generation e.g.\ for news. \citet{tan2017neural} use a coarse-to-fine approach which first identifies important sentences and then converts them into a headline. In this way, the model is not confused by `too much' irrelevant information. In A2T, the first summarization step 
may not be necessary, as the abstract is already a summary of the scientific paper. 
}

How titles should be (and are) structured has been researched for a long time, e.g., 
\cite{lewison2005s}. \citet{hartley2008academic} gives a typology of title types, distinguishing 13 title classes, e.g., those that state results vs.\ methods.  

Beyond title generation, related fields of text generation for science are related work generation \citep{li2022generating}, \serev{more general automatic paper section writing assistance \citep{wang-etal-2019-paperrobot}},    
and 
automatically generating reviews for scientific articles \citep{yuan2022can}. 
\sen{More broadly relating to science, Meta has in 2022 released an LLM for the scientific domain called Galactica \citep{Taylor2022GalacticaAL}, but they mostly explore it for scientific classification tasks rather than generation.}

\paragraph{Humor identification and generation}
Humor detection is a niche area in  NLP but nonetheless with a rich history. For example, 
\citet{mihalcea2006learning} distinguish funny from non-funny sentences (heuristically scraped from the web) using features and traditional classifiers. \citet{simpson2019predicting} focus on efficiently annotating humor and inducing classifiers from crowd-sourced data. 
\citet{PeyrardBG021} show that transformers are strong at distinguishing funny from non-funny sentences on minimal pairs of satirical news headlines. In the scientific domain, \citet{Heard2022IfTT} annotate a dataset of more than 2k titles from ecology using a fine-grained Likert scale. The majority were labeled as non-funny and annotators exhibited low agreements. \sen{ 
\citet{shani-etal-2021-get} classify scientific titles as funny or not using humor-theory inspired features and scientific language models such as SciBERT \citep{beltagy-etal-2019-scibert} building on a dataset of Ig Nobel winners and humorous papers discussed in online forums.}

There is considerably less work on humor generation. As one exception, \citet{he-etal-2019-pun} generate puns by a retrieve-and-edit approach based on word2vec, thus circumventing the problem of little training data for puns. 
\section{Data}
\label{sec:data}

We use the dataset released by \citet{DetectingStanceInScientificPapers}, 
which 
contains title-abstract pairs and corresponding meta-information such as 
publication year and venue. \citet{DetectingStanceInScientificPapers} extracted the data from two sources: ACL Anthology (from 1984 to 2021) and machine learning conferences (from 1989 to 2021); we refer to the datasets from these two sources as \texttt{NLP} and \texttt{ML}, respectively. 
After filtering \serev{(described in Appendix \ref{app:filter})},
\textbf{32,952} abstract-title pairs remain in our dataset.

\section{Title Generation} 
We first explore whether 
existing state-of-the-art Seq2Seq models
manage to generate human-level titles from abstracts. Hence, we do not include humor constraints. 
\ycrev{We use an 8:2 ratio to divide the data into train and test sets, and randomly select 1,000 instances from the train set for the dev set.}

\subsection{Models}

We experiment with the following six generation models: 
(i) BART base (\texttt{BART$_\text{base}$}) \citep{lewis-etal-2020-bart}, (ii) GPT2  (\texttt{GPT2}) \citep{radford2019language}, (iii) T5 small \citep{2020t5} (\texttt{T5}), and (iv) PEGASUS large \citep{zhang2019pegasus} finetuned on Extreme Summarization (XSUM) dataset \citep{xsum-emnlp} (\texttt{PEGASUS$_\text{xsum}$}). Noting the similarity between text summarization and our A2T generation task, we additionally inspect two BART large models finetuned on (v) XSUM  (\texttt{BART$_{\text{xsum}}$}) and (vi) CNN dailymail (CNNDM) \citep{see-etal-2017-get} (\texttt{BART$_{\text{cnn}}$}), respectively. 
XSUM and CNNDM 
contain document-summary pairs, where XSUM has one-sentence summaries, while each summary in CNNDM consists of multiple sentences. 

\paragraph{Fine-tuning} \label{sec:finetune_baselines}
For all baseline models, we continue fine-tuning them on the abstract-title pairs from our 
dataset. Details are in Appendix \ref{app:train_title}. 

\subsection{Evaluation}\label{sec:gen1_eval}
We assess the performance of the 
systems on 230 abstracts 
using both automatic evaluation metrics and human evaluation. 
We also include the human-generated titles in the evaluation, 
denoted as `\texttt{HUMAN}'. \serev{While our test set is small, we note that (i) human evaluation is very time-consuming and (ii) we have more source-output pairs (i.e., 230$\times$6, see below) than in some standard MT or summarization evaluation benchmarks such as WMT15-17 or SummEval \citep{Fabbri2020SummEvalRS}.}

\paragraph{Automatic Evaluation:} As there are no A2T task-specific evaluation metrics, we use the following 
metrics 
from other NLG tasks: 
Rouge \citep{lin2004rouge}, 
BERTScore \citep{bert-score}, MoverScore \citep{zhao-etal-2019-moverscore}, COMET \citep{rei-etal-2020-comet}, BARTScore \citep{yuan2021bartscore}, MENLI \citep{menli}. 
\se{COMET is a metric supervised on human scores from 
MT, 
all others 
are unsupervised.}
We employ all 
metrics in both \emph{reference-based} and \emph{-free} settings. 
Reference-based, 
the metrics compare the system titles with the original 
human-generated titles, 
while 
reference-free, 
the system titles are directly compared to the abstracts. 
The details of the metric variants can be found in Appendix \ref{app:variant_metrics}.
The reference-free setup is more consistent with our human evaluation below and overall more plausible for A2T. 

\paragraph{Human Evaluation:}
The human evaluation is conducted 
reference-free: 
15 annotators\footnote{Most annotators are Master students, with an additional senior researcher and two Bachelor students.} were asked to select two best and two worst titles among six titles from different systems (including \texttt{HUMAN}), given the abstract. 
\sen{In order to make the annotation simpler for humans, we only considered one dimension of annotation, namely, `overall quality', which may comprise aspects such as fluency, (grammatical) correctness, adequacy, etc. 
This mimics coarse-grained annotations such as direct assessment (DA) in fields like MT. We did not further subdivide the quality into more fine-grained subcategories, as the annotation is already difficult and comprises to understand a scientific abstract and to decide which title best fits it.}
Each instance (an abstract and its six titles) was evaluated by 
\serev{at least two annotators; depending on availability, some instances were annotated by up to five annotators.}
The average percentage agreement over all annotator pairs is $\sim$50\%, implying that each two annotators 
agree on one selection among the two selected best/worst titles, on average. 

Then, we use best-worst scaling (\textbf{BWS}) \citep{louviere1991best} to obtain the final human score for each title as:  
\begin{align}
    \textit{BWS} &= \frac{N_{\textit{best}}-N_{\textit{worst}}}{N_{\textit{annotators}}}
\end{align}
where $N_{\textit{best/worst}}$ refers to the number of times that the title was selected as one of the best/worst two titles and $N_{annotators}$ indicates the number of annotators responsible for that instance. 

\begin{table}[!htb]
\resizebox{\columnwidth}{!}{%
\begin{tabular}{@{}lccccccc@{}}
\toprule
system        & BWS  & MoverS & BERTS & BARTS & COMET       & MENLI   & ROUGE    \\ \midrule
\texttt{BART$_{\text{xsum}}$} & {\ul \textbf{0.197}} & -0.025               & {\ul \textbf{0.889}  }              & {\ul \textbf{-2.583}} & \textbf{0.060} & -0.214   & 0.033       \\
\texttt{PEGASUS$_\text{xsum}$} & 0.022  & -0.036     & 0.887     & -2.819    & 0.060       & -0.263   & 0.035    \\
\texttt{BART$_\text{base}$}    & 0.015  & -0.034     & 0.887     & -2.709    & 0.059       & -0.226    & 0.035   \\
\texttt{GPT2}          & -0.013 & -0.087     & 0.881     & -3.090    & 0.060       & -0.285   & 0.020     \\
\texttt{T5}    & -0.039 & -0.055     & 0.889     & -2.735    & 0.057       & -0.265    & 0.032   \\
\texttt{BART$_{\text{cnn}}$}  & -0.384               & {\ul \textbf{0.046}} & 0.880 & -2.982                & 0.047          & \textbf{-0.159}  & {\ul \textbf{0.055}} \\ \midrule
\texttt{HUMAN}         & 0.181  & -0.062     & 0.873     & -3.508    & {\ul 0.061} & {\ul -0.029} & 0.029 \\ \bottomrule
\end{tabular}%
}
\caption{Ref-free evaluation results of the baseline models. We underlie the best performance among all generation systems including human. We bold the best performance among all automatic generation systems excluding human.}
\label{tab:eval_baseline}
\end{table}

\paragraph{Results}
 We present the \textbf{reference-based evaluation} results 
 in Appendix \ref{app:ref_based_baseline}. 
\emph{Among the six 
systems, \texttt{BART$_{\text{xsum}}$} is best}, being selected by \yc{4 out of 6} evaluation metrics, followed by \texttt{BART$_{\text{cnn}}$}.

\begin{figure}[!t]
    \centering
    \includegraphics[width=\columnwidth]{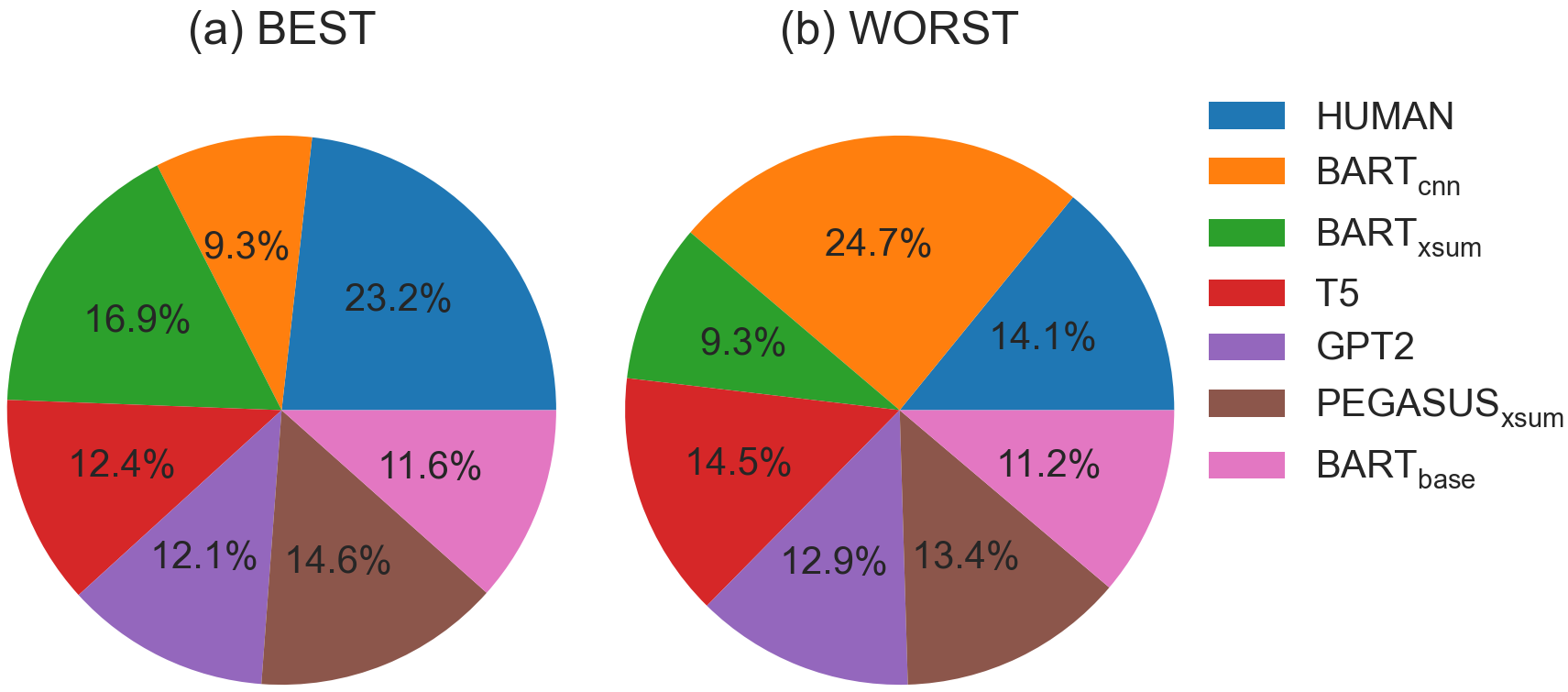}
    
    \caption{Distribution of generation systems of the titles selected as the \emph{BEST}/\emph{WORST} ones in human evaluation; percentages indicate the proportion of the generation systems being selected over all selections.}
    \label{fig:best_worst_distribution}
\end{figure}

Table \ref{tab:eval_baseline} shows the \textbf{reference-free evaluation} 
results. 
Unlike in 
reference-based evaluation, only two evaluation metrics (COMET
and MENLI) select \texttt{HUMAN} as the best 
system. \texttt{BART$_{\text{xsum}}$} is still the best among the six automatic 
systems, obtaining best results on 4 out of \yc{7} evaluation metrics (including BWS). Surprisingly, it outperforms \texttt{HUMAN} even in the human evaluation (0.197 vs.\ 0.181 BWS). Nevertheless, as Figure \ref{fig:best_worst_distribution}(a) shows, \texttt{HUMAN} was still most frequently selected as among the two best titles (23.2\%) among all generation systems, whereas the best neural generation system \texttt{BART$_{\text{xsum}}$} was selected in 16.9\% of the cases as one of the best two titles. 
However, 
Figure \ref{fig:best_worst_distribution}(b) 
shows 
that \texttt{HUMAN} was also more often selected as among the two worst titles (14.1\% vs.\ 9.3\% \texttt{BART$_{\text{xsum}}$}), 
explaining why 
\texttt{BART$_{\text{xsum}}$} is better than \texttt{HUMAN} in human evaluation. 
Introspection shows that this is mostly due to words in the human title which do not appear in the abstract. As a consequence, human annotators may believe that the model is hallucinating. 
Overall, we thus believe that there is a (slight) mismatch in our task definition: human authors may leverage the whole paper when designing their titles, not only the abstracts. However, paper2title generation would not only be a challenge for the text generation models (which are often limited in text length) but also for the human annotation process. We argue that framing the problem as abstract2title generation is a simplification with overall good tradeoffs between problem complexity and model and annotator capacity. 

\paragraph{Why is the best model best?}
\ycrev{To get a deeper insight into the quality of the system titles, we first analyze their lengths. 
\texttt{BART$_{\text{cnn}}$} produces titles much longer than human titles  
(14.95 vs.\ 8.27 tokens) and 
other 
systems (6.68-9.13 tokens), on average; besides, its titles are often truncated due to the maximal output length set to the model. This reflects the mismatch of the training data---\texttt{BART$_{\text{cnn}}$} was first trained on CNNDM which has multiple sentences as a summary.
Among the other systems, \texttt{BART$_{\text{xsum}}$} and \texttt{BART$_{\text{base}}$} generate titles having the largest overlap with the abstracts, based on the edit distance. 
} 
\serev{While \texttt{BART$_{\text{xsum}}$} (best/worst: 241/133) does not have a huge advantage over \texttt{BART$_{\text{base}}$} (best/worst: 165/159), inspection of results indicates that  \texttt{BART$_{\text{xsum}}$} may give more precise and relevant titles, e.g., it picks out the key information from the abstracts more frequently; some examples are in Appendix \ref{app:bartbase}. This may be 
due to its (extreme) summarization objective in the pre-training phase.} 
 
\subsection{
Reliability of Evaluation Metrics}
\label{sec:auto_metric}

To inspect the reliability of the used metrics, we calculate \yc{Spearman/}Pearson correlation with system-level human judgments, i.e., average BWS per system, on the 1380 titles (230 instances $\times$ 6 titles). 
From Table \ref{tab:metric_correlation} (left block), we observe: (1) most metrics perform better in the ref-based setup than ref-free, except for COMET. (2) Only ref-free COMET correlates well with human judgments from the perspective of both types of correlation. 

Even though COMET 
performs well on system-level,
this 
only indicates that COMET ranks systems 
similarly 
as humans.  
COMET is not necessarily 
good at selecting the best title among different choices 
(segment-level evaluation). Indeed, 
at segment-level, 
\yc{it correlates weakly with human scores (0.127 Kendall)}.\footnote{As we convert BWS to WMT relative ranking judgements \citep{ma-etal-2018-results}, we use the Kendall-like formulation introduced there for segment-level correlation.}
Inspired by this, we train a ref-free metric supervised on our own human scores.

\subsection{A2TMetric}\label{app:a2tmetric}

\begin{table}[!t]
\centering
\resizebox{\columnwidth}{!}{%
\begin{tabular}{@{}l|cccc|cc@{}}
\toprule
          & \multicolumn{4}{c|}{230 instances}                                 & \multicolumn{2}{c}{35 instances}                                            \\
          & \multicolumn{2}{c}{ref-based}   & \multicolumn{2}{c|}{ref-free}    & \multicolumn{2}{c}{ref-free}                                                \\
          & $\rho$       & $r$       & $\rho$       & $r$       & $\rho$                             & $r$                              \\ \midrule
ROUGE     & 0.571          & 0.395          & -0.250         & -0.722         & -0.121$\pm$0.11         & -0.404$\pm$0.26         \\
BARTS     & 0.393          & 0.389          & 0.214          & -0.044         & 0.200$\pm$0.30          & 0.083$\pm$0.21          \\
BERTS     & 0.571          & 0.442          & 0.250          & 0.079          & 0.236$\pm$0.26          & 0.296$\pm$0.22          \\
MoverS    & 0.929          & 0.575          & -0.071         & -0.677         & -0.129$\pm$0.13         & -0.378$\pm$0.24         \\
MENLI     & 0.357          & 0.345          & 0.321          & 0.139          & 0.057$\pm$0.15          & 0.160$\pm$0.21          \\
COMET     & \textbf{0.964} & \textbf{0.580} & \textbf{0.929} & \textbf{0.929} & 0.414$\pm$0.32          & 0.679$\pm$0.15          \\ \midrule
A2TMetric & -              & -              & -              & -              & \textbf{0.707$\pm$0.17} & \textbf{0.726$\pm$0.16} \\ \bottomrule
\end{tabular}%
}
\caption{Pearson's $r$ 
and Spearman's $\rho$ of evaluation metrics with \textbf{system-level} human judgements for all \textbf{230 instances} (1380 titles; left block) 
and \textbf{35 instances} (210 titles; right block). The correlations on the 35 instances are averaged over the test sets from five splits.
We bold the highest correlation in each block.}
\label{tab:metric_correlation}
\end{table}

We develop the first \textbf{supervised A2T generation-specific evaluation metric}, using the human judgments collected in the evaluation for the 230 instances. Since 
\yc{\texttt{HUMAN}} as a generation system 
is included in the evaluation, and the metrics will later be used to evaluate system-generated humorous titles, which may vastly differ from the original ones, we argue that
a ref-free metric will better suit our needs. 

\paragraph{Dataset} We split the data of 230 instances to train (170 instances), dev (25 instances), and test (35 instances) sets. 
\yc{To get more robust results, we generate five different splits of train, dev and test sets and report the average performance of the metrics on the test sets over the five splits in Table \ref{tab:metric_correlation}.}
We note that many titles receive a BWS of 0 when the number of annotators is small (because they were never selected as the best or worst two titles), which may be problematic 
when aiming to 
directly train a regression model. Besides, the human evaluation was similar to the ranking process.  
Therefore,
we convert BWS in the train and dev set to relative-ranking judgments \citep{ma-etal-2018-results}.  
That is, 
if two titles for one abstract obtain different BWS, this title pair is considered as one relative-ranking judgement. Each 
instance then contains one abstract, a ``better'' title, a ``worse'' title, and the score difference between the two titles in addition. 

\paragraph{Framework} We adopt a 
framework 
similar 
to the ranking-based variant of COMET to train the A2T metrics but in a ref-free setup. During training, the model optimizes the embedding space so that (1) the sentence embedding of the abstract ($a$) is closer to that of the ``better'' title ($t^{+}$) than to that of the ``worse'' title ($t^{-}$) (using the Triplet Margin loss \citep{7298682}) and (2) the difference between $d(a,t^{+})$ and $d(a,t^{-})$ is close to the difference in BWS human scores for the two titles (using the MSE loss), where $d(u,v)$ refers to the Euclidean distance between $u$ and $v$. During predicting, the metrics calculate the Euclidean distance between the sentence embeddings of the abstract and the title.

\paragraph{Evaluation} 
\yc{As Table \ref{tab:metric_correlation} (right block) shows, our A2TMetric
achieves the highest values of both average Spearman and Pearson correlations (above 0.71-0.73 vs.\ -0.40-0.68) and relatively low standard deviation (around 0.16 vs.\ 0.11-0.32), implying that it is not only superior to the existing metrics but also demonstrates comparably good robustness. 
}
 
\rmd{
While the metric is still not of absolutely high quality segment level (0.276 Kendall), 
it clearly outperforms COMET and the other metrics 
(right half of Table \ref{tab:seglevel_metrics} in the appendix) and the correlation values are on the same level as those of the best MT metrics in WMT22 shared Task \citep{freitag-etal-2022-results}. 
System-level, we evaluate A2TMetric on 5 random samples of size 35 where the remainder instances are for train/dev. While 
there is a high variance due to small sample size, A2TMetric is on average 0.1-0.3 Pearson/Spearman better 
system-level 
than COMET (right block of Table \ref{tab:metric_correlation}).  
Even though comparing the trained A2TMetric to 
unsupervised metrics 
may seem unfair, 
this is exactly the key point: A2TMetric is better because it has been trained on our costly human data, which makes it valuable.
}

COMET is still the best among the existing metrics.
Therefore, we only leverage our trained A2TMetric
and COMET to automatically evaluate the A2T
systems’ quality in 
\S\ref{sec:humor_eval}. 

\section{Humorous Title Generation}\label{sec:humor}
\sen{To generate humorous titles, we first need a dataset of humor annotated titles in our domain (NLP and ML papers). We cannot resort to the data of \citet{shani-etal-2021-get,Heard2022IfTT} as those leverage papers from other scientific fields. As a consequence, we build our own dataset. 
When constructing the dataset, we ask annotators to rely on their intuition of humor rather than issuing guidelines of what they should find funny. This can be justified as humor is often subjective and culture- and even gender-specific \citep{dore2019humour,mundorf1988gender}. 
There is also a multitude of theories around humor, indicating the ambiguity of the concept.\footnote{The wikipedia page for humor \url{https://en.wikipedia.org/wiki/Theories_of_humor} lists at least three modern popular theories of humor, based on relief, superiority and incongruity.}
}

\paragraph{Humor Annotation + Classification}
We train humor classifiers \sen{on human annotated data} to automatically label 
titles as \funny{}, \medfunny{}, and \nfunny{} (examples see Table \ref{tab:huomr_example} the appendix). 
Two 
co-authors  
participated in the annotation. 
\sen{Examples of their annotations are shown in 
\ycn{Appendix \ref{app:funny_annotated_titles}}. Titles annotated as funny by both annotators allude to famous proverbs or book/movie titles (``\emph{Taming the wild}''), make use of linguistic devices such as alliteration (``\emph{Balancing Between Bagging and Bumping}'') or leverage surprise (``\emph{Is the Best Better? [...]''; ``What's in a name? In some languages, grammatical gender}''). Medium funny titles often make use of playful/clever abbreviations, e.g., ``\emph{CPR: Classifier-Projection Regularization for Continual Learning}'' --- such playful abbreviations seem characteristic especially for the NLP community (BERT, LLAMA, etc.).} 

\textbf{Stage 1}:   
The two annotators initially annotated \textbf{1,730} titles: 
1,603 titles 
as \nfunny{}, 106 as \medfunny{}, and 21 as \funny{} (kappa 0.65 on 300 common instances). 
\sen{To combat this severe data imbalance, we resort to ensembling with each classifier trained on more balanced splits:} 
we randomly generate 11 different data splits, where the train set of each split consists of 100 funny or medium funny titles and 200 not funny titles (all randomly drawn).  
On those 
splits, we train 11 classifiers to construct an ensemble classifier. 
To evaluate the classifier performance, the two annotators annotated another 315 titles jointly, obtaining 0.639 Kappa. 
Our best ensemble classifier leverages the sum of the label values assigned by the 11 individual classifiers to predict 
humorousness, yielding 4.8\% macro F1 improvement 
compared to the individual classifiers (62.4\% vs. 57.6\%).
Details are in Appendix \ref{app:humor_anno}.

\textbf{Stage 2}:
To find more funny title candidates to annotate, the two annotators annotated the funniest 396 titles in the original dataset from \citet{DetectingStanceInScientificPapers}, 
predicted by the Stage 1 ensemble classifier;   
75.8\% (300 titles) were \se{judged} 
as \funny{} or \medfunny{}, which is substantially higher than the proportion of funny titles in the annotated data of Stage 1 (7.3\%). 
Thus, the annotated data expands to 
\textbf{2,441} 
titles
($= 1,730+315+396$),
where 1,893 are labeled as \nfunny{}, 492 as \medfunny{} and 56 as \funny{}. Subsequently, we re-train 11 classifiers on 
newly generated 11 data splits from the expanded data of 2,441 titles; now the train set of each split 
has 
400  
(medium) funny 
titles 
and 800 not funny titles. As before, 
\serev{we ensemble the 11 classifiers}
\serev{as}
in Stage 1.

\begin{table}[!t]
\centering
{%
\begin{tabular}{@{}lcc@{}}
\toprule
                         & Individuals            & Ensemble \\ \midrule
Stage 1 &  52.2 / 81.5  & 54.1 / 85.1\\ \midrule
Stage 2 &  55.1 / 84.7  & \textbf{57.7} / \textbf{88.1} \\ \bottomrule
\end{tabular}%
}
\caption{Average macro F1 over the 11 individual classifiers and macro F1 of the ensemble classifiers from both stages on the held-out test set (where the two annotators obtain 0.649 kappa agreement). \ycrev{Performance on both three-way (first entry) and binary (second entry) classification tasks; for \emph{binary} classification, \funny{} and \medfunny{} 
are merged.}
We bold the highest macro F1 on each classification task.}
\label{tab:eval_two_stages}
\end{table}

We test the classifiers from both stages on a held-out test set containing 197 titles
annotated by the two annotators (0.649 kappa). 
The macro F1 scores of those classifiers are presented in Table \ref{tab:eval_two_stages}. As \funny{} titles are rare in the whole dataset,
we also evaluate the classifiers on the corresponding binary classification task where \funny{} and \medfunny{} are merged. 
We observe that: (1) the ensemble classifier performs better than the individual ones. (2) Classifiers from Stage 2 are superior 
to the ones from Stage 1, indicating 
larger size of the training data is beneficial.  
(3) The best three-way classifier achieves only $\sim$58\% macro F1,  
but 
$\sim$88\% macro F1 
on the 
binary classification. 
Besides, we see a consistent improvement of human annotation quality: the two annotators achieve 0.01-0.1 higher Kappa 
when their annotations are down-scaled to binary 
(see Table \ref{tab:kappa} in Appendix \ref{app:humor_anno}).  
\textbf{Thus, we use the ensemble classifier from Stage 2 as the humor classifier in further experiments}.

\paragraph{Final Dataset}
We use our humor classifier to automatically label the rest of the data. 
Considering the difficulty of
three-way classification 
for both humans 
and 
classifiers, we only consider two humor levels in further 
experiments: (1) \funny{} (for funny and medium funny titles) and (2) \nfunny{} (for not funny titles).
Thus, we collect 31,541 instances ($>$95\%) with \nfunny{} and 1,411 with \funny{} titles. 
We split the resulting data to train, dev, and test sets, ensuring that (1) the data with human-annotated titles remains in the train set, 
as the humor classifier trained and evaluated on it will be used as an automatic humor evaluator; (2) 80\% of the data in dev/test is from \texttt{NLP} and 20\% from \texttt{ML} because our annotators are more knowledgable for NLP papers, and (3) the ratio of \funny{} data to \nfunny{} data \yc{in dev/test set} is 1:2.\footnote{
This aims to more easily compare the system-generated funny titles with the human-generated ones and does not relate to controlling the quality of titles in the test set.}
As \funny{} data is only a small portion of the whole data, we only keep 600 instances in the dev/test sets, 
the remaining data serves as the train data. \ycn{Appendix \ref{app:data_statistic}}
summarizes the statistics of the final dataset.

\paragraph{Generation} In the second phase of the experiments, we use the optimal model identified previously, i.e., \texttt{BART$_{\text{xsum}}$}, to generate titles with constraints on humor level. The input of the generation systems is formulated as ``\emph{humor level [SEP] abstract}'', where humor level is either 0 (for {\nfunny{}}) or 1 (for \funny{}).

\paragraph{Fine-tuning}
We fine-tune generation systems here  
as in 
\S\ref{sec:finetune_baselines} (\ycn{hyperparameters see Appendix \ref{app:para_train_2}}): (1) we fine-tune a \texttt{BART$_{\text{xsum}}$} on the abstract-title pairs in the train set with humor constraints.
(2) We continue fine-tuning the model from (1) on self-generated pseudo data.\footnote{\serev{Synthetic data can be a useful resource 
\citep{He2021GenerateAA}, despite potential limitations \citep{Shumailov2023TheCO}.}}

The motivation of (2) is that we observe that the systems tend to ignore the humor constraints in the input and generate identical titles for different constraints in initial experiments. We assume that to expose 
systems 
to titles with different humor levels for the same abstract 
during training can encourage them 
to pay more attention to the humor constraints. To obtain the pseudo data, we: 
\serev{(i) generate titles for abstracts in the train set but with ``opposite'' humor constraints compared to the original titles, keeping only those pseudo titles with the correct humor labels assigned by the humor classifier; (ii) filter out \funny{} labeled titles with very frequent n-grams, in order to encourage more diverse titles. We finally merge the filtered pseudo data with the original data. Thus, in the training data of (2), each abstract has two titles, 
    one with label \funny{} and the other with {\nfunny{}};
\ycrev{it contains 15,474 instances in total, where 50\% are pseudo ones.}}

\subsection{Evaluation}
\label{sec:humor_eval}
\ycrev{We report results on generating both funny and not-funny titles, to explore the difference in models' performance after involving humor generation, based on both automatic and human evaluation.}

\paragraph{Automatic Evaluation}
Based on the 
results for the automatic evaluation metrics in  \S\ref{sec:auto_metric}, we only leverage \textbf{COMET} and our supervised metric \textbf{A2TMetric} 
here to evaluate 
title 
quality. To evaluate 
\serev{humor}, 
we use the following three metrics: (1) \textbf{F1$_{\text{macro}}$} between the expected humor labels and those assigned by the humor classifier. (2) System accuracy of generating titles on correct humor levels, denoted as \textbf{ACC$_{\text{FUNNY}}$} and \textbf{ACC$_{\neg{\text{FUNNY}}}$}. (3) The ratio of the cases that the systems generate the same titles for both humor constraints to all generation cases (\textbf{Ratio$_{\text{SAME}}$}); lower 
is better. 

We generate titles with constraint on both humor levels for all abstracts in the test set,  
computing the automatic evaluation on 1200 titles in total.

\begin{table}[]
\centering
\resizebox{\columnwidth}{!}{%
\begin{tabular}{@{}lcc|cl@{}}
\toprule
Metric            & \multicolumn{2}{c|}{COMET}  & \multicolumn{2}{c}{A2TMetric} \\
humor constraint  & $\neg{\text{FUNNY}}$        & FUNNY       & $\neg{\text{FUNNY}}$         & FUNNY         \\ \midrule
\texttt{BART$_{\text{xsum}}$}        & \textbf{0.0598}        & 0.0582      & \textbf{-2.30}       & -2.32       \\
\texttt{BART$_{\text{xsum}}$+pseudo} & 0.0593        & 0.0541      & -2.31       & -2.37       \\ \midrule
\texttt{HUMAN}             & \multicolumn{2}{c|}{0.0586} & \multicolumn{2}{c}{-2.36}   \\ \bottomrule
\end{tabular}%
}
\caption{Automatic evaluation for titles' quality. We bold the best performance assessed by each metric.  ``Humor constraint'' refers to the constraints given to the input of the generation systems.}
\label{tab:auto_eval_quality_humor}
\end{table}

\paragraph{Results}
\yc{We evaluate 
humor before and after training on pseudo data in Appendix \ref{app:auto_eval_humor}}, \serev{Table \ref{tab:auto_eval_humor}}:
(1) after continued training on the pseudo data, 
\texttt{BART$_{\text{xsum}}$+pseudo} achieves substantially higher F1$_{\text{macro}}$ (from 0.647 to 0.856) and ACC$_{\funny{}}$ (from 40.2\% to 77.8\%), and slightly better Ratio$_{\text{SAME}}$ (from 6.5\% to 4.7\%). 
(2)
ACC$_{\neg\emph{FUNNY}}$ drops slightly compared to \texttt{BART$_{\text{xsum}}$} (94.5\% vs.\ 93.6\%), indicating that both systems have
high accuracy on generating \nfunny{} titles and
the fine-tuning on pseudo data only 
improves the system's 
accuracy to generate \funny{} titles. 

We then present the quality 
evaluation results in Table \ref{tab:auto_eval_quality_humor}.  
Both BART 
systems 
obtain better results than \texttt{HUMAN} on both evaluation metrics, which is in line with the observation in  \S\ref{sec:gen1_eval}, especially when generating \emph{$\neg{\text{FUNNY}}$} titles. 
However, we observe a consistent performance drop after training on the pseudo data (values in the first row vs.\ those in the second row). Further, we also note that the system generated \emph{$\neg{\text{FUNNY}}$} titles have better quality than the \funny{} ones (values in the left column vs.\ those in the right column).  

\paragraph{Human Evaluation}\label{humor_human_eval}
We randomly sample \yc{100} abstracts from the test set with controls on the source of the papers (80\% from \texttt{NLP} and 20\% from \texttt{ML})  
and on the humor label of the original titles (50\% \funny{} and 50\% {\nfunny{}}). 
\se{For each abstract with a human funny title, we generate a funny and a non-funny system title, and accordingly for each non-funny human title.} 
\yc{
Thus, each evaluation instance contains one abstract and five titles: 1 original title + 4 system titles (2 generation systems $\times$ 2 humor levels). 
The annotators rank the five titles on two criteria: \emph{general quality} and \emph{humor degree}, based on the abstract; 
the annotators can assign identical ranks to multiple titles. 
\yc{We show a screenshot of an annotation instance and the annotation guidelines in Figure \ref{fig:anno_example} in the appendix.}
Five annotators (three PhD students, one undergraduate student and one senior researcher) jointly annotate 10 from these 100 instances, obtaining 0.782 Spearman for humor and 0.325 for quality ranking on average per annotator pair. 
Then, they separately evaluate the remaining 90 instances. 
} 
\sen{Note that since in our evaluation annotators rank titles, even the first ranked title does not necessarily have to be of high quality or funny, for any given abstract, if the remaining are very bad concerning quality/humor.}

\begin{table}[!t]
\centering
\resizebox{\columnwidth}{!}{%
\begin{tabular}{@{}lcccc@{}}
\toprule
humor constraint/label & \multicolumn{2}{c}{\funny{}} & \multicolumn{2}{c}{\emph{$\neg${FUNNY}}} \\ 
system                 & humor      & quality      & humor       & quality       \\ \midrule
\texttt{BART$_{\text{xsum}}$}             & 1.94       & \textbf{2.70}         & 2.76        & \textbf{2.10}          \\
\texttt{BART$_{\text{xsum}}$+pseudo}     & 1.58       & 2.97         & 2.75        & 2.56          \\\midrule
\texttt{HUMAN}                  & \textbf{1.51}       & 2.86         & \textbf{2.40}        & 2.63          \\ \bottomrule
\end{tabular}%
}
\caption{Average rank of the system titles for the abstracts with original titles labeled as \funny{} and \nfunny{} separately in the human evaluation of general quality and humor degree; smaller values denotes higher ranks. ``Humor constraint/label'' refers to the constraints given to the input of the generation systems and the humor labels of the original titles.}
\label{tab:human_eval_humor_sep}
\end{table}

\paragraph{Results}
Table \ref{tab:human_eval_humor_all} (appendix)  
compares the two BART 
systems \se{across all 200 instances (one funny and one non-funny title per abstract)}. Similar to automatic evaluation, we observe (1) a general quality drop but a performance boost for humor generation after training on 
pseudo data and (2)  {\nfunny{}} titles have better quality than \funny{} ones.

Further, we compare the system titles with the original human 
titles in Table \ref{tab:human_eval_humor_sep}. 
\yc{
\texttt{BART$_\text{xsum}$} 
ranks higher than \texttt{HUMAN} concerning quality when generating both \funny{} and \nfunny{} titles (2.70 vs.\ 2.86 and 2.10 vs.\ 2.63), which is consistent with our previous human evaluation (\S\ref{sec:gen1_eval}).  
However, fine-tuning on the pseudo data impacts the quality of the generated funny titles, as the system is rated worse than \texttt{HUMAN} only in this category (2.97 vs.\ 2.86), which is also in line with our automatic evaluation from A2TMetric.
\texttt{HUMAN} still generates funnier titles than the automatic 
systems, ranking highest among all systems (1.51 vs.\ 1.58-1.94).
}

\section{Comparison with ChatGPT} \label{sec:chatgpt}
\begin{table}[!t]
\centering

\resizebox{0.7\columnwidth}{!}{%
\begin{tabular}{@{}lcc@{}}
\toprule
system   & humor rank & quality rank \\ \midrule
\texttt{BART$_{\text{xsum}}$} & 1.86 / 2.66  & 2.74 / \textbf{2.25}    \\
\texttt{ChatGPT}  & \textbf{1.41} / 3.12  & 3.62 / 2.30    \\ \midrule
human    & 2.53       & 2.85         \\ \bottomrule
\end{tabular}}
\caption{Average ranks of the generated \funny{} titles (first entry) and \nfunny{} titels (second entry) for \textbf{100 abstracts from EMNLP 2022 handbook} in the human evaluation of quality and humorousness; smaller values denote higher ranks. We bold the highest ranks for each criterion.}
\label{tab:chatgpt_1}
\end{table}

We compare our fine-tuned \texttt{BART$_{\text{xsum}}$} (without training on pseudo data) with the recent popular ChatGPT  
model.\footnote{\ycrev{Here, we used the ChatGPT interface (\url{https://chat.openai.com/}) of the first three releases (Nov. 30, 2022---Jan. 9, 2023); the official API was inaccessible back then.}} 
Firstly, we use the two models to generate funny and not funny titles for \yc{100} abstracts from the 
EMNLP 2022 handbook 
which ChatGPT could not have seen in its training data. 
\sen{Our prompt} 
for ChatGPT is \emph{``I want a funny title and a not-funny title for the following abstract: [abstract]''}. 
The ranking-based human evaluation conducted here is identical to  
\S\ref{humor_human_eval} \yc{and done by the same five annotators, who obtain 0.867 Spearman for humor and 0.548 for quality evaluation on average over annotator pairs this time}.  

The average rank per system with humor constraint is presented in Table \ref{tab:chatgpt_1}. We observe that 
\yc
{automatic generation systems are mostly ranked higher than \texttt{HUMAN} (2.25-2.74 vs.\ 2.85) except for \texttt{ChatGPT} producing funny titles (3.62 vs.\ 2.85). 
\texttt{ChatGPT} generates funnier but lower-quality titles compared to \texttt{BART$_{\text{xsum}}$} \se{but ChatGPT is almost on par for non-funny titles}. 
Hence, we conclude that \emph{\texttt{ChatGPT} without any fine-tuning may already perform similarly to our fine-tuned \texttt{BART$_{\text{xsum}}$}}.
}

\ycrev{After our experiments, ChatGPT has been updated several times. 
To inspect whether the new version performs better, 
we conduct a second experiment 
using the \serev{latest} model ``gpt-3.5-turbo-0613'' with the official API, utilizing the default hyperparameters. 
\serev{Details are given in Appendix \ref{app:chat_version}}. 
\ycrev{Overall, our evaluation suggests that \emph{the newer ChatGPT does not perform better:} 
In 25 out of 40 cases, the previous titles were selected as the better ones. 
In fact, the new version performs much worse for generating \funny{} titles: 
it loses to the previous \serev{version} on 18 out of 20 instances.}
}
\section{Discussion \& Analysis}

\paragraph{Are automatic titles really superior?} Overall, our results 
in \S\ref{sec:humor} and \S\ref{sec:chatgpt}
seem to indicate that automatically generated titles outperform human 
titles.  
However, 
looking at the distribution of best/worst titles, we see again a high frequency of worst human titles as annotated by our human annotators; in fact, human titles are most frequently selected as worst titles except when the automatic systems use the humor constraint. As before, the likely reason is 
a lower lexical overlap between human titles and abstracts. 
\se{Indeed, we find that human titles have lower lexical overlap with abstracts when compared to automatically generated titles from \texttt{ChatGPT} and \texttt{BART$_\text{xsum}$}, 
e.g., \yc{57-61\%} of content words in human titles appear in the abstract, while the number is \yc{64-67\%} for \texttt{BART$_\text{xsum}$} and \texttt{ChatGPT}.} 
 \se{Very negatively evaluated human titles have even lower lexical overlap.}

In contrast, 
human titles were again most frequently selected as best titles except 
when 
including ChatGPT. 
Overall, our findings implicate that automatically generated titles can be competitive but are presumably still slightly worse than author choices. To verify this hypothesis, \emph{we suggest a more costly evaluation scheme in the form of a user study involving the authors of papers instead of paper external annotators in future studies.}

\paragraph{Is training on extra parts besides abstract beneficial?}
\ycrev{
\serev{We argued that human titles may not only be based on abstracts, but (to some extent) the full papers. To inspect whether training title generation systems on more than abstracts alone leads to better systems, we }
train \texttt{BART$_\text{xsum}$}  and \serev{the popular} Longformer-Encoder-Decoder (\texttt{LED} ) \citep{beltagy2020longformer}, \serev{which can deal with longer input sequences,} in two settings: (1) only 
abstracts and (2) 
abstracts, introductions, and conclusions; we denote the corresponding models as ``\texttt{[MODEL]+A}'' and ``\texttt{[MODEL]+X}'', respectively. 
We use the data from \citet{hou-etal-2021-tdmsci}, which contains the sentences of all papers from ACL Anthology until 2019. \serev{Technical details are given in Appendix \ref{app:extra_train}. }}

\ycrev{
We randomly select 29 instances from the test sets for human evaluation: 
14 
for \texttt{BART$_\text{xsum}$}  and 15 
for \texttt{LED} . 
Two evaluators were asked to select the better one among the two titles generated by ``\texttt{[MODEL]+A}'' and ``\texttt{[MODEL]+X}'' with the same underlying model, given the abstract, introduction and conclusion. 
On the jointly assessed 10 instances, they obtained 0.474 Kappa. 
}
%
\ycrev{Our evaluation results show that: \texttt{BART$_\text{xsum}$}  seems to benefit from training on more parts (\texttt{BART$_\text{xsum}$+X} wins 8 out of 14 instances); for \texttt{LED} , it is not the case (\texttt{LED+A} wins 11 out of 15 cases).
\serev{On introspection, we do find that the models trained on more than abstracts can indeed 
leverage 
some relevant keywords not in the abstracts, which makes their titles sometimes better. 
On the other hand, they are tasked with identifying relevant titles given more `background noise' (longer texts) which causes them to hallucinate more and be more vague. 
We show examples in Appendix \ref{app:ax}. 
Evaluation with more than abstracts alone is also considerably more costly for humans. Overall, these experiments thus indicate that training (and evaluating) on highly specific and condensed abstracts is advantageous.}
}

\paragraph{Humor constraints} On introspection, we find that the funny titles generated by ChatGPT do not conform to a style of 
humor used in scientific papers. 
This indicates that \emph{ChatGPT lacks fine-tuning on humor in science}. 
For {\texttt{BART$_{\text{xsum}}$}, its problem 
seems to be that it overfits to data artefacts learned from the data 
\emph{indicating that it does not properly learn a generalizable notion of humor}. Additionally, both models often do not match the content of the abstract/title to the humor framing \yc{(examples see Table \ref{tab:example_humor_constraints} in the appendix).} 
\ycn{In our human evaluation, such titles often obtain high humor but low quality ranks; however, when they are pertinent to the abstracts, they have the potential to receive high quality ranks as well (cf.\ Appendix \ref{app:low_quality}).}  

\section{Conclusion}
We considered the abstract-to-title generation problem using end-to-end models. To do so, we trained six recent text-to-text generation systems on more than 30k NLP and ML papers. We evaluated the systems using an array of state-of-the-art automatic metrics as well as human evaluation. Our evaluation indicates that some current text generation models can generate titles with similar quality as humans, but human authors are apparently still superior. We also considered the humorous title generation problem as an extension, compiling the first dataset in the NLP/ML domain in this context, comprising over 2.6k titles annotated by two annotators with acceptable agreement. We find that our systems struggle with generating humorous titles and instead overfit to frequent patterns in the data, indicating much scope for future research.  
\section{Limitations}
In our work, we followed a standard protocol of evaluation of text generation involving (1) automatic metrics comparing source texts (abstracts) or references to system outputs and (2) human annotators considering the same sources of information. We argued that this standard evaluation scheme may not be fully adequate in our situation as the human authored titles may take additional information into account (e.g., the full texts), which is difficult to incorporate, however, for our annotators and for the metrics. 
This leads to an (arguably small) bias against human titles, which seems to be automatically identifiable however via the distribution of best/worst titles selected for different systems. 
Overall, this 
limitation 
could better be addressed, however, by consulting the authors of papers for an additional \sen{but much} more costly to realize evaluation in the form of a user study. 

We also 
experimented with 
NLP and ML papers only, not taking other scientific fields into consideration. 
Finally, prompting for ChatGPT is an art in itself; other prompts may have yielded different results. 
To explore this, 
we used a slightly different prompt (``\emph{Please give me a [funny] title for the following scientific abstract: [abstract]}'') for ChatGPT on 20 instances, which led to very similar human evaluation results. It is conceivable, however, that there might have been prompts leading to  better evaluation outcomes for ChatGPT. 

A risk of our models is that 
they might produce misleading or even factually wrong titles 
which could be adopted by the human authors if not properly checked. 

\sen{As a consequence of our missing annotation guidelines for humor, it is possible that our annotators have not clearly separated humor from related concepts such as `click-baiting' (to the extent that such a separation is possible at all).}

\section*{Acknowledgments}
We 
thank the BMBF for its support via the grant Metrics4NLG. The last author is supported by DFG grant
EG 375/5–1.

\bibliography{anthology,custom}
\bibliographystyle{acl_natbib}

\appendix

\section{Filtering}
\label{app:filter}
(1) {We restrict the data to the main conference papers} (e.g., EMNLP, ACL). 
{We limit the data to abstracts of length smaller than 400 words} 
 as extremely long abstracts \yc{in the dataset}
often contain extra sections other than abstracts. 
{(3) We only leverage 
papers published after the year 2000} (which form the majority anyway). 

\section{Training details for title generation}
\label{app:train_title}
We train models with AdamW Optimizer \citep{loshchilov2018decoupled} and linear learning rate scheduler, and subsequently use beam search \citep{vijayakumar2016diverse} as the sampling strategy to generate the output candidates. The optimal checkpoint for each model is selected based on the ROUGE1/2/L \citep{lin2004rouge} scores on the dev set. 
Table \ref{tab:hyp_title_generation} displays the hyperparameter 
for training 
and Table \ref{tab:beam_title_generation} shows the parameters used for beam search. The models were trained using Google Colab with a Tesla K80 GPU which has 24 GB of memory. We show the number of parameters of each baseline model in Table \ref{tab:num_parameters}.

\begin{table*}[!htb]
\centering
\resizebox{.8\textwidth}{!}{%
\begin{tabular}{@{}lcccc@{}}
\toprule
             & learning rate & batch size & epochs & gradient accumulation steps \\ \midrule
\texttt{BART$_{\text{xsum}}$}   & 3e-05         & 3          & 3      & 8                           \\
\texttt{PEGASUS$_\text{xsum}$} & 6e-04         & 3          & 3      & 8                           \\
\texttt{BART$_\text{base}$}    & 3e-04         & 8          & 3      & 8                           \\
\texttt{GPT2}         & 3e-04         & 2          & 3      & 8                           \\
\texttt{T5}     & 3e-04         & 8          & 3      & 8                           \\
\texttt{BART$_{\text{cnn}}$}     & 3e-04         & 4          & 3      & 8                           \\ \bottomrule
\end{tabular}%
}
\caption{Training hyperparameter for title generation. We use the AdamW optimizer with a weight decay of 0.01 and keep the other settings as default in Huggingface's Trainer API.}
\label{tab:hyp_title_generation}

\vspace{1cm}
\resizebox{0.25\textwidth}{!}{%
\begin{tabular}{@{}lc@{}}

\toprule
max length           & 30 \\
min length           & 3  \\
repetition penalty   & 2  \\
length penalty       & 10 \\
num beams            & 5  \\
num return sequences & 5  \\\bottomrule
\end{tabular}%
}
\caption{Parameter settings for beam search.}
\label{tab:beam_title_generation}

\end{table*}

\section{Variants of used automatic evaluation metrics}\label{app:variant_metrics}
In ref-based evaluation, we report Rouge-1 recall, BERTScore recall, unigram MoverScore, BARTScore recall, MENLI(ref$\leftarrow$cand\_e-c) 
and COMET(wmt20-comet-da). In ref-free setup, we use the Faithfulness variant for BARTScore,  MENLI(src$\rightarrow$cand\_-c) and COMET (wmt21-comet-qe-mqm) instead; the variants of the other metrics are the same as in ref-based setting.

\section{Ref-based evaluation results of baseline models}\label{app:ref_based_baseline}
Table \ref{tab:eval_baseline_old} shows the ref-based automatic evaluation results of the baseline models.
\begin{table}[!htb]
\centering
\resizebox{\columnwidth}{!}{%
\begin{tabular}{@{}lcccccc@{}}
\toprule
system        & MoverS     & BERTS      & COMET           & BARTS       & MENLI   & ROUGE    \\ \midrule
\texttt{BART$_{\text{xsum}}$}    & \textbf{0.410} & \textbf{0.912} & \textbf{-0.283} & -3.816          & 0.076  & \textbf{0.455}     \\
\texttt{PEGASUS$_\text{xsum}$} & 0.404          & 0.906          & -0.371          & -3.964          & 0.005      & 0.384 \\
\texttt{BART$_\text{base}$}    & 0.405          & 0.907          & -0.373          & -3.986          & 0.036     & 0.403  \\
\texttt{GPT2}          & 0.400          & 0.902          & -0.461          & -4.114          & -0.020    & 0.361  \\
\texttt{T5}     & 0.381          & 0.898          & -0.501          & -4.177          & -0.025   & 0.337   \\
\texttt{BART$_{\text{cnn}}$}     & 0.282          & 0.907          & -0.634          & \textbf{-3.747} & \textbf{0.133}     & 0.448  \\ \bottomrule
\end{tabular}%
}
\caption{Ref-based evaluation results of the baseline models. We underlie the best performance among all generation systems including human. We bold the best performance among all automatic generation systems excluding human.}
\label{tab:eval_baseline_old}
\end{table}

\section{BART$_\text{base}$ VS.\ BART$_{\text{xsum}}$}\label{app:bartbase}
Table \ref{tab:bartbase} shows the examples of abstract-title pairs where \texttt{BART$_\text{base}$} failed to capture the key information in the abstract while \texttt{BART$_{\text{xsum}}$} succeeded.
\begin{table*}[!hbt]
\centering
\small
\begin{tabularx}{\textwidth}{lX}
\toprule
\texttt{Abstract}   & [...] we propose to learn word embeddings based on the recent \colorbox{y}{fixed-size} ordinally forgetting encoding (FOFE) method, which can almost uniquely encode any variable-length sequence into a \colorbox{y}{fixed-size} representation. [...]  \citep{sanu-etal-2017-word}                                                                                                                                      \\ \midrule
\texttt{BART$_\text{base}$} & Learning Word Embeddings Based on Ordinally Forgetting Encoding                                                                                                                                                                                                                                                                                                              \\
\texttt{BART$_{\text{xsum}}$} & Learning Word Embeddings Based on \colorbox{green}{Fixed-Size} Ordinally Forgetting Encoding                                                                                                                                                                                                                                                                                                   \\ \midrule\\ \midrule
                                                                   
\texttt{Abstract}   & [...] Unfortunately, the reliance on manual annotations, which are both difficult and highly expensive to produce, presents a major obstacle to the widespread application of these systems across different languages and text genres. In this paper we describe a method for inducing the semantic roles of verbal arguments directly \colorbox{y}{from unannotated text}. [...] \citep{lang-lapata-2010-unsupervised} \\ \midrule
\texttt{BART$_\text{base}$} & Inducing Semantic Roles from Text for Semantic Role Labeling                                                                                                                                                                                                                                                                                                                 \\
\texttt{BART$_{\text{xsum}}$} & A Probabilistic Model for Semantic Role Induction \colorbox{green}{from Unannotated Text}                                                                                                                                                                                                                                                                                                      \\ \midrule \\ \midrule

\texttt{Abstract}    & [...] At the same time, we argue that \colorbox{y}{relation labeling} can benefit from naked tree structure and should be treated elaborately with consideration of three kinds of relations including within-sentence, across-sentence and across-paragraph relations. Thus, we design a pipelined two-stage parsing method for generating an \colorbox{y}{RST} tree from text. [...] \citep{wang-etal-2017-two}         \\ \midrule
\texttt{BART$_\text{base}$} & Pipelined Two-Stage Parsing of Named Discourse Trees                                                                                                                                                                                                                                                                                                                         \\
\texttt{BART$_{\text{xsum}}$} & Pipeline-based Parsing of Discourse Trees for \colorbox{green}{RST} and \colorbox{green}{Relation Labeling}                                                                                                                                                                                                                                                                                                      \\ \bottomrule
\end{tabularx}
\caption{Examples of abstract-title pairs where \texttt{BART$_\text{base}$} failed to capture the key information in the abstract while \texttt{BART$_{\text{xsum}}$} succeeded. The key information is highlighted in both abstracts and titles.}
\label{tab:bartbase}
\end{table*}

\begin{table}[!htb]
\centering
\begin{tabular}{lc|c}
\hline
          & 230 instance & 35 instances \\
          & $\tau$          & $\tau$          \\ \hline
ROUGE     & -0.054       & -0.014       \\
BARTS     & 0.092        & 0.121        \\
BERTS     & 0.078        & 0.113        \\
MoverS    & 0.001        & 0.038        \\
MENLI     & 0.061        & 0.121        \\
COMET     & \textbf{0.127}        & 0.194        \\ \hline
A2TMetric & -            & \textbf{0.276}        \\ \hline
\end{tabular}
\caption{Segment-level WMT $\tau$-like correlations of ref-free evaluation metrics on all 230
instances (1380 titles; left block) and 35 instances (210
titles; right block). The correlations on the 35 instances
are averaged over the test sets from five splits. We bold
the highest correlation in each block.}
\label{tab:seglevel_metrics}
\end{table}

\begin{table}[!htb]
\centering
\vspace{-.2cm}
\resizebox{\columnwidth}{!}{%
\footnotesize
\noindent\begin{tabularx}{\columnwidth}{ >{\hsize=.77\hsize}X | >{\hsize=.23\hsize}X }
\toprule
  Title &
  Label \\ \midrule
  Learning to learn by gradient descent by gradient descent \citep{andrychowicz2016learning} &
  \funny{} \\ \midrule
  CancerEmo: A Dataset for Fine-Grained Emotion Detection \citep{sosea-caragea-2020-canceremo} & \medfunny{} \\ \midrule
  Global Encoding for Abstractive Summarization \citep{lin-etal-2018-global} & \nfunny{} \\ \bottomrule
\end{tabularx}%
}
\vspace{-.2cm}
\caption{Examples of annotated titles. }
\label{tab:huomr_example}
\vspace{-.4cm}
\end{table}

\section{Examples of funny titles}\label{app:funny_annotated_titles}

Table \ref{table:most_funny} and Table \ref{tab:example_funny_titles} show sample funny titles labeled by human annotators. 
\sen{We note: some instances of humor 
require contextual (e.g., culture- or domain-specific) knowledge such as references to popular TV shows (`Germany's next language model'); 
this is characteristic of humor and 
makes it challenging/subjective. Despite of this, our agreements indicate a shared notion of humor among our annotators.}

\begin{table*}[!hb]
\small
\centering
\resizebox{\textwidth}{!}{%
\begin{tabularx}{\textwidth}{ >{\hsize=\hsize}X}
\toprule
Towards Multimodal Sarcasm Detection (An \_Obviously\_ Perfect Paper) \citep{castro-etal-2019-towards}                                                           \\
Thieves on Sesame Street! Model Extraction of BERT-based APIs \citep{Krishna2019ThievesOS}                                                                   \\
Are Two Heads Better than One? Crowdsourced Translation via a Two-Step Collaboration of Non-Professional Translators and Editors \citep{yan-etal-2014-two}\\
Taming the Wild: A Unified Analysis of Hogwild-Style Algorithms \citep{desa2015taming}                                                                  \\
Balancing Between Bagging and Bumping \citep{NIPS1996_f4733064}                                                                                           \\
Speculation and Negation: Rules, Rankers, and the Role of Syntax \citep{velldal-etal-2012-speculation}                                                                 \\
What's in a name? In some languages, grammatical gender \citep{nastase-popescu-2009-whats}                                                                          \\
BAM! Born-Again Multi-Task Networks for Natural Language Understanding \citep{clark-etal-2019-bam}                                                           \\
Is the Best Better? Bayesian Statistical Model Comparison for Natural Language Processing \citep{szymanski-gorman-2020-best}                                       \\
Keep CALM and Explore: Language Models for Action Generation in Text-based Games \citep{yao-etal-2020-keep}                                                 \\ \bottomrule
\end{tabularx}%
}
\caption{Examples of \textbf{human} titles which were labeled as \medfunny{}+\medfunny{}, \medfunny{}+\funny{}, or \funny{}+\funny{}
by the two annotators (the two entries denote the label assigned by different annotators.).}
\label{table:most_funny}
\end{table*}

\begin{table*}[!hb]
\small
\centering
\resizebox{\textwidth}{!}{%
\begin{tabularx}{\textwidth}{ >{\hsize=\hsize}X}
\toprule
\funny{}  \\                                                                                           \midrule
German's Next Language Model \citep{chan-etal-2020-germans} \\
Is the Best Better? Bayesian Statistical Model Comparison for Natural Language Processing \citep{szymanski-gorman-2020-best} \\
Comparing Apples to Apple: The Effects of Stemmers on Topic Models \citep{schofield-mimno-2016-comparing} \\
(Almost) No Label No Cry \citep{NIPS2014_a8baa565}                                                                    \\
The Trumpiest Trump? Identifying a Subject's Most Characteristic Tweets \citep{pethe-skiena-2019-trumpiest}                              \\
Questionable Answers in Question Answering Research: Reproducibility and Variability of Published Results \citep{crane-2018-questionable}   \\
Know What You Don't Know: Unanswerable Questions for SQuAD \citep{rajpurkar-etal-2018-know}                                         \\
Dear Sir or Madam, May I Introduce the GYAFC Dataset: Corpus, Benchmarks and Metrics for Formality Style Transfer \citep{rao-tetreault-2018-dear} \\
Can You Tell Me How to Get Past Sesame Street? Sentence-Level Pretraining Beyond Language Modeling \citep{wang-etal-2019-tell} \\
Showing Your Work Doesn't Always Work \citep{tang-etal-2020-showing}\\
"Got You!": Automatic Vandalism Detection in Wikipedia with Web-based Shallow Syntactic-Semantic Modeling \citep{wang-mckeown-2010-got}\\
It's a Contradiction - no, it's not: A Case Study using Functional Relations \citep{ritter-etal-2008-contradiction}\\
\midrule
\medfunny{}                                                                                                   \\ \midrule
CPR: Classifier-Projection Regularization for Continual Learning \citep{DBLP:journals/corr/abs-2006-07326} \\
NYTWIT: A Dataset of Novel Words in the New York Times \citep{pinter-etal-2020-nytwit} \\
MedDialog: Large-scale Medical Dialogue Datasets \citep{zeng-etal-2020-meddialog}                                                           \\
Catching Captain Jack: Efficient Time and Space Dependent Patrols to Combat Oil-Siphoning in International Waters \citep{10.5555/3504035.3504061}         \\
The Shattered Gradients Problem: If resnets are the answer, then what is the question? \citep{shattered}                         \\
Go Simple and Pre-Train on Domain-Specific Corpora: On the Role of Training Data for Text Classification \citep{edwards-etal-2020-go}                      \\
SentiLARE: Sentiment-Aware Language Representation Learning with Linguistic Knowledge \citep{ke-etal-2020-sentilare}                   \\ 
Get Semantic With Me! The Usefulness of Different Feature Types for Short-Answer Grading \citep{pado-2016-get}\\
Witches' Brew: Industrial Scale Data Poisoning via Gradient Matching \citep{https://doi.org/10.48550/arxiv.2009.02276}\\
ENGINE: Energy-Based Inference Networks for Non-Autoregressive Machine Translation \citep{tu-etal-2020-engine}\\
You Can't Beat Frequency (Unless You Use Linguistic Knowledge) - A Qualitative Evaluation of Association Measures for Collocation and Term Extraction \citep{wermter-hahn-2006-cant}\\
OntoGUM: Evaluating Contextualized SOTA Coreference Resolution on 12 More Genres \citep{zhu-etal-2021-ontogum}\\
\bottomrule
\end{tabularx}%
}
\caption{Selected 
\textbf{human} titles in the annotated data \se{judged as funny or medium funny by the annotators}.}
\label{tab:example_funny_titles}
\end{table*}

\begin{table}[!htb]
\centering
\resizebox{.65\columnwidth}{!}{%
\begin{tabular}{@{}lc@{}}
\toprule
             & \# parameters \\ \midrule
BART$_{\text{base}}$    & 140M          \\
BART$_{\text{xsum}}$    & 400M          \\
BART$_{\text{cnn}}$     & 400M          \\
T5    & 60M           \\
GPT2         & 117M          \\
PAGASUS$_{\text{xsum}}$ & 568M          \\ \bottomrule
\end{tabular}
}
\caption{Number of parameters of the six baseline models.}
\label{tab:num_parameters}
\end{table}

\section{Humor annotation + classifiation}\label{app:humor_anno}

The two annotators first annotated the same \textbf{230 titles} independently, obtaining only {0.397 Kappa} agreement, which indicates a relatively bad annotation quality. To improve the inter-agreement between the annotators, they 
then discussed the reasons leading to 
disagreement. 
Subsequently, they annotated another \textbf{300 titles} independently, achieving a decent {0.650 Kappa} 
for a task as subjective as humor. 
As a consequence, \emph{we use the maximal label value among the two annotations for each title as its final label for the 300 titles}, i.e., if one annotator labels a title with 1 (\medfunny{}), while the other labels with 0 (\nfunny{}), we assign label 1 to the title.
Each annotator then labeled 600 different titles separately, bringing \textbf{1,730} ($230+300+600\times2=1730$) 
annotated titles in total, 
where 1,603 titles are labeled as \nfunny{}, 106 as \medfunny{} and 21 as \funny{}. 

As the funny titles (labeled as \funny{}) are very few compared to the not funny ones (labeled with 0), we generate 11 different data splits, where the train set of each split consists of 100 funny titles and 200 not funny ones (randomly sampled from the 1730 titles), while the remaining 27 funny titles and other 27 not funny ones compose the dev set. 
From the 11 different data splits, we obtain 11 classifiers (checkpoints selected based on the macro F1 on each dev set). 
We then evaluate the ensembles of the 11 classifiers on \textbf{315 newly annotated titles} by the two annotators, who obtain \textbf{0.639 Kappa} agreement this time. With this step, we study the optimal ensemble of the classifiers  
and also obtain more funny titles from the whole data by annotating the funniest titles selected by the ensemble classifiers. We design two types of ensemble classifiers: 
\begin{itemize}
    \item \textbf{EnsMV}, which relies on the majority vote of the 11 classifiers. Specifically, each title receives 11 labels from the 11 classifiers: if the number of \nfunny{} labels exceeds 5, 
    the title is labeled as \nfunny{}; if not, the title is labeled as \funny{} when the number of \funny{} labels exceeds the number of \medfunny{} labels, otherwise it is labeled as \medfunny{}.
    
    \item \textbf{EnsSUM$_{i,j}$}, which depends on the sum of the label values. The sum of the label values for each title ranges from 0 (11 classifiers $\times$ 0 for \nfunny{}) to 22 (11 classifiers $\times$ 2 for \funny{}). We then select a threshold $i$ for \medfunny{} and $j$ for \funny{}: if sum $< i$, the title is labeled as \nfunny{}; otherwise it is labeled as 
    \medfunny{} (when sum $< j$) or \funny{} (when sum $\ge j$).
\end{itemize}

\begin{table}[!hb]
\centering
\resizebox{0.8\columnwidth}{!}{%
\begin{tabular}{@{}lccc@{}}
\toprule
   & \multirow{2}{*}{Individuals} & \multicolumn{2}{c}{Ensembles} \\
   &                              & EnsMV   & EnsSUM$_{\text{7,16}}$  \\ \midrule
F1 & 57.6\%                       & 61.4\%          & \textbf{62.4\%}      \\ \bottomrule
\end{tabular}%
}
\caption{Average macro F1 over the 11 individual classifiers and macro F1 of the ensemble classifiers from stage 1 on the evaluation data of 315 titles (where the two annotators obtain 0.639 kappa). We bold the highest macro F1 score.}
\label{tab:eval_ensemble}

\end{table}

Table \ref{tab:eval_ensemble} shows the evaluation results of Stage 1; we only present the performance of EnsSUM$_{i,j}$ with optimal $i$ and $j$ here, i.e., EnsSUM$_{7,16}$. We observe that: (1) both ensembles perform better than the individual ones (+4-5\% macro F1) and (2) \textbf{EnsSUM$_{\text{7,16}}$} is slightly better than EnsMV (62.4\% vs. 61.4\% macro F1).

\begin{table}[!hb]
\centering
\resizebox{0.7\columnwidth}{!}{%
\begin{tabular}{@{}cccc@{}}
\toprule
        &          & \multicolumn{2}{c}{Kappa}  \\
        & \#titles & three-way & binary         \\ \midrule
\multirow{3}{*}{Stage 1} & 230      & 0.397     & \textbf{0.513} \\
        & 300      & 0.650     & \textbf{0.754} \\
        & 315      & 0.639     & \textbf{0.709} \\ \midrule
Stage 2 & 197      & 0.649     & \textbf{0.661} \\ \bottomrule
\end{tabular}%
}
\caption{Kappa agreements between the two annotators on several data pieces. ``\#titles'' refers to the number of titles in a certain piece of data. We bold the higher Kappa on the same data.}
\label{tab:kappa}
\end{table}

\section{Dataset Statistics}\label{app:data_statistic}
Table \ref{tab:dataset_distribution} shows the statistics of the final dataset.
\begin{table}[!htb]
\centering
\resizebox{\columnwidth}{!}{%
\begin{tabular}{@{}lcc|c|cc@{}}
\toprule
      & \multicolumn{2}{c|}{Humor label} & \multirow{2}{*}{Total} & \multicolumn{2}{c}{Source} \\
      & $\neg$\emph{FUNNY}         & FUNNY          &                        & NLP          & ML          \\ \midrule
train & 30,741           & 1,011           & 31,752                  & 16,141        & 15,611       \\
dev   & 400             & 200            & 600                    & 480          & 120         \\
test  & 400             & 200            & 600                    & 480          & 120         \\
total & 31,541           & 1,411           & 32,952                  & 17,101        & 15,851       \\ \bottomrule
\end{tabular}%
}
\caption{Distribution of the source (NLP or ML) and humor labels (\funny{} or \nfunny{}) of the instances in our dataset.}
\label{tab:dataset_distribution}
\end{table}

\section{Parameters for humor generation}\label{app:para_train_2}
We train \texttt{BART$_{\text{xsum}}$} on our train set using the AdamW optimizer with weight decay 0.01 and learning rate 4e-05 for 5 epochs. Then we continue to train it on the pseudo data for one epoch to obtain \texttt{BART$_{\text{xsum}}$+pseudo}. We use the default settings in Huggingface's Trainer API for the other hyperparameters. We train the models with an RTX A6000 GPU
which has 48 GB of memory.

To monitor the models' ability to generate titles on correct humor levels, we use \emph{macro F1} between the expected humor labels (i.e., the humor constraints given to the inputs) and the humor labels assigned to the generated titles by the humor classifier as the performance indicator, with which on the dev set we select the optimal model checkpoints of the two systems.

\section{Automatic evaluation of humor generation}\label{app:auto_eval_humor}
Table \ref{tab:auto_eval_humor} shows the systems' ability for humor generation before and after training on the pseudo data according to the automatic evaluation.
\begin{table}[!htb]
\centering
\resizebox{\columnwidth}{!}{%
\begin{tabular}{@{}lcccc@{}}
\toprule
                  & F1$_{\text{macro}}$       & ACC$_{\neg\emph{FUNNY}}$ & ACC$_{\funny{}}$ & Ratio$_{\text{SAME}}$    \\ \midrule
\texttt{BART$_{\text{xsum}}$}        & 0.647          & \textbf{94.5\%} & 40.2\%           & 6.5\%          \\
\texttt{BART$_{\text{xsum}}$+pseudo} & \textbf{0.856}$\uparrow$ & 93.6\%$\downarrow$          & \textbf{77.8\%}$\uparrow$  & \textbf{4.7\%}$\uparrow$ \\ \bottomrule
\end{tabular}%
}
\caption{Automatic evaluation for the systems' ability to generate titles with correct humor constraints. We bold the best performance. $\uparrow$/$\downarrow$ in the second row indicates the performance being better/worse after training on the pseudo data.}
\label{tab:auto_eval_humor}
\end{table}

\begin{table}[!h]
\centering
\resizebox{\columnwidth}{!}{%
\begin{tabular}{@{}lccc@{}}
\toprule
system                             & humor constraint & humor & quality \\ \midrule
\multirow{2}{*}{\texttt{BART$_{\text{xsum}}$}}        & \nfunny{}     & 2.85  & \textbf{2.32}    \\
                                   & \funny{}       & 1.79  & 2.81    \\ \midrule
\multirow{2}{*}{\texttt{BART$_{\text{xsum}}$+pseudo}} & {\nfunny{}}     & 2.97  & 2.64    \\
                                   & \funny{}      & \textbf{1.43}  & 3.26    \\ \bottomrule
\end{tabular}%
}
\caption{Average rank of the system titles for all abstracts in the human evaluation of general quality and humor degree; smaller values denotes higher ranks. ``Humor constraint'' refers to the constraints given to the input of the generation systems.}
\label{tab:human_eval_humor_all}
\vspace{-.3cm}
\end{table}

\section{Examples of system-generated funny 
titles}\label{app:low_quality}
Table \ref{tab:example_low_quality_funny_titles} and \ref{tab:example_high_quality_funny_titles} show 10 system-generated low-quality funny titles and 10 system-generated high-quality funny titles, respectively, according to the human evaluation results.

\begin{table*}[!htb]
\centering
\small
\resizebox{\textwidth}{!}{%
\begin{tabularx}{\textwidth}{ >{\hsize=\hsize}X}
\toprule 
\texttt{BART$_{\text{xsum}}$} - funny titles with artefacts                                                                             \\ \midrule
\emph{What's in} a Semantic Model? Comparing LDA and LSA on the Web \citep{10.5555/2390948.2391052} \\
\emph{Don't} paraphrase unless you know what you are talking about: Improving Question Answering Performance by Paraphrasing \citep{Duboue2006AnsweringTQ}  \\
\emph{Don't} Transliterate, Use Context! Mining New Word Translations from Comparable Corpora Using Context Information \citep{li2004integrating}\\
\emph{Reading Between the Lines:} Unsupervised Summarization of Spontaneous Speech using Acoustic Patterns \citep{10.5555/1690219.1690223}\\

 \midrule
\texttt{ChatGPT} - non-scientific funny titles                                                                                  \\ \midrule
Proof Generation: Now You See It, Now You Don't! \citep{https://doi.org/10.48550/arxiv.2205.12443} \\
Co-Guiding Net: Helping You Hit the Slot and Intent Jackpot!  \citep{https://doi.org/10.48550/arxiv.2210.10375} \\
Abduct Me If You Can: How to Prove a Claim With a Little Help From Your Friends (Premises) \citep{https://doi.org/10.48550/arxiv.2211.00614}\\
OREO-LM: The Creamy, Crunchy, and Smart Way to Answering Open-Domain Questions \citep{https://doi.org/10.48550/arxiv.2211.08380} \\
 \bottomrule
\end{tabularx}%
}
\caption{Examples of system-generated funny titles from \texttt{BART$_\text{xsum}$} with artefacts and non-scientific funny titles from \texttt{ChatGPT}. The citations here are the original papers for those titles.}
\label{tab:example_humor_constraints}
\end{table*}

\begin{table*}[!htb]
\centering
\small
\resizebox{\textwidth}{!}{%
\begin{tabularx}{\textwidth}{ >{\hsize=\hsize}X}
\toprule
\texttt{BART$_{\text{xsum}}$}                                                                                \\ \midrule
Don't Invite Adversaries to Poison Your Data: Exploiting Federated Learning for Adversarial Backdoor Attacks \citep{yoo-kwak-2022-backdoor}                   \\
Don't Take the Easy Way Out: Generating Adversarial Negative Responses with Large-Scale Language Models for Dialogue Selection \citep{lee-etal-2022-pneg} \\
Don't Give Up on Style: Learn to Generate Stylistically-Diverse Summaries with Multiple Decoders \citep{goyal-etal-2022-hydrasum}                              \\
CKD: Curriculum Knowledge Distiller for Cross-Lingual Sentiment Analysis with Emoji \citep{zhang-etal-2022-curriculum}      \\
Successive Prompting: Learning to Break Down Complex Questions into As Simple As Possible \citep{dua-etal-2022-successive}\\ \midrule
\texttt{ChatGPT}                                                                                  \\ \midrule
Graphin' It Up: A Humorous Guide to Generative Knowledge Construction \citep{ye-etal-2022-generative}                    \\
Tiny Tasks, Big Results: A Hilarious Guide to Few-Shot Relation Extraction \citep{li-qian-2022-graph}               \\
Revealing the Magic Behind Transformer Language Models: A Lighthearted Investigation \citep{geva-etal-2022-transformer}     \\
Ask and You Shall Receive: A Whimsical Approach to Automatic Question Generation \citep{wang-etal-2022-learning-generate}         \\
Federated Learning: The More You Poison, the More You Win! \citep{yoo-kwak-2022-backdoor}                                \\ \bottomrule
\end{tabularx}%
}
\caption{Examples of system-generated \textbf{low-quality} funny titles, which obtain high humor ranks but low quality ranks in the human evaluation.}
\label{tab:example_low_quality_funny_titles}
\end{table*}

\begin{table*}[!htb]
\centering
\small
\resizebox{\textwidth}{!}{%
\begin{tabularx}{\textwidth}{ >{\hsize=\hsize}X}
\toprule
\texttt{BART$_{\text{xsum}}$}                                                                                \\ \midrule
Don't Agree with Me? Introducing Semantic Environment Features Improves Agreement-Disagreement Classification in Online Discourse \citep{gokcen-de-marneffe-2015-disagree}                   \\
The Myth of the Two Sides of the Same Coin: Claim Generation and Claim Retrieval in a World of Claims \citep{gretz-etal-2020-workweek}\\
Sharing is Caring: Incentives for Self-Organization in Social Welfare Maximization \citep{Gollapudi_Kollias_Panigrahi_2019}                             \\
DeCEMBERT: Dense Captions and Entropy Minimization for Video-and-Language Pre-training \citep{tang-etal-2021-decembert}     \\
Stochastic Alternating Direction Method of Multipliers Revisited: Faster Rates and Better Algorithms \citep{pmlr-v32-azadi14}\\ \midrule
\texttt{ChatGPT}                                                                                  \\ \midrule
Succeed with Successive Prompting: Breaking Down Complex Questions for LMs \citep{dua-etal-2022-successive}                   \\
Feeling the Pulse of Dialogue: A Supervised Prototypical Contrastive Learning for Emotion Recognition in Conversation \citep{song-etal-2022-supervised}               \\
Triple Trouble: A Novel Query-Based Approach to Joint Entity and Relations Extraction \citep{tan-etal-2022-query}     \\
Two Heads are Better than One: A Multi-View Fusion and Multi-Decoding Method for Multi-Document Reading Comprehension \citep{wen-etal-2022-m3}          \\
Seeing is Believing: A Picture's Worth a Thousand Words in Multimodal Machine Translation \citep{ji-etal-2022-increasing}                               \\ \bottomrule
\end{tabularx}%
}
\caption{Examples of system-generated \textbf{high-quality} funny titles, which obtain both high humor and quality ranks in the human evaluation.}
\label{tab:example_high_quality_funny_titles}
\end{table*}

\begin{figure*}[h]
    \centering
    \includegraphics[width=\textwidth]{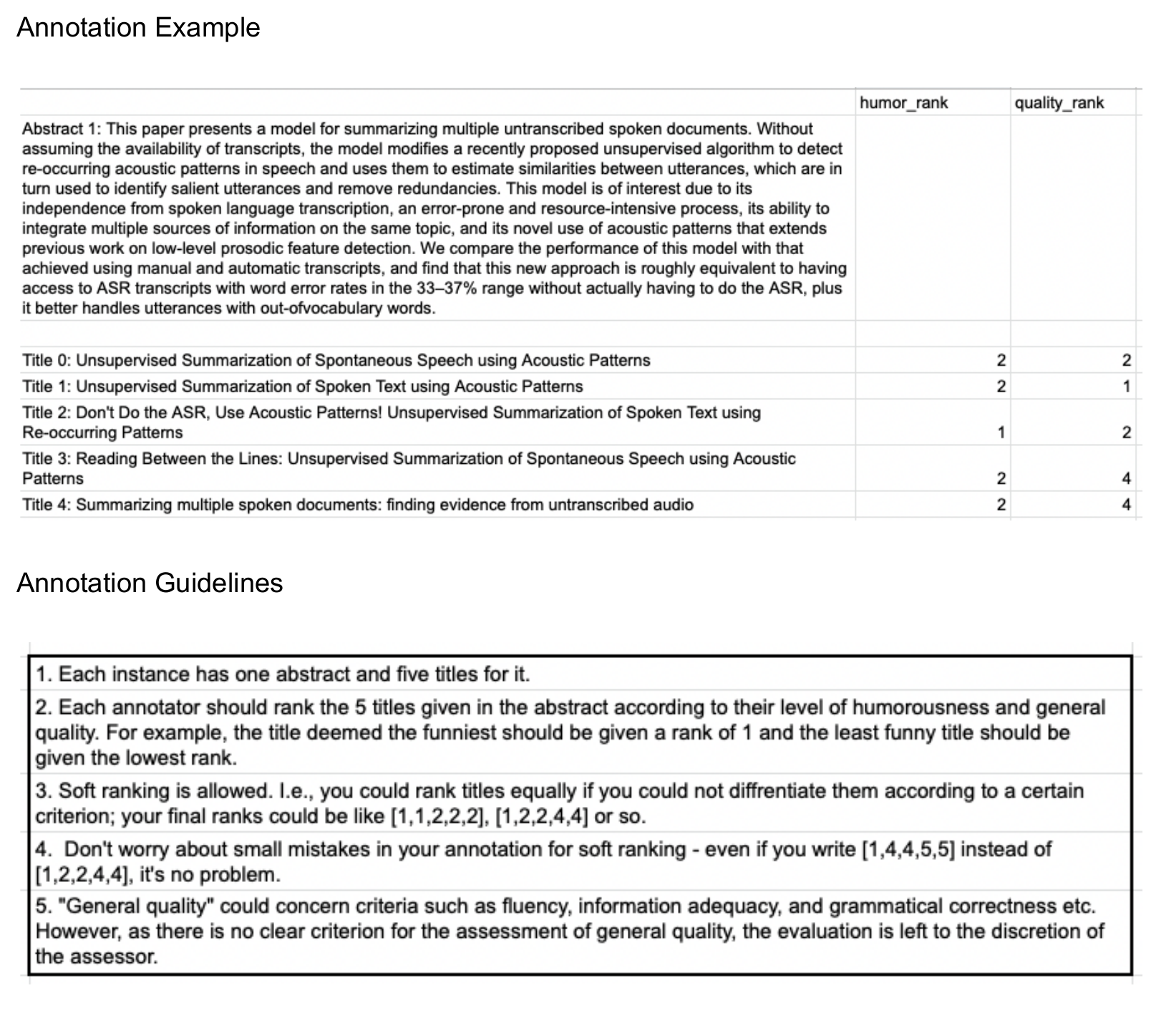}
    \caption{Screenshot of an annotation instance and the annotation guidelines. The evaluation is conducted with google spreadsheet.}
    \label{fig:anno_example}
\end{figure*}

\section{Comparison of ChatGPT versions}
\label{app:chat_version}

\ycrev{We randomly choose 10 abstract-title pairs from our previous evaluation for both low- and high-quality titles, following each humor constraint (\funny{} and \nfunny{});
this totals to 40 evaluation instances.\footnote{In this context, we consider the titles ranked above 2 as high quality and below 3 as low quality.}
Then, we use the new version of ChatGPT to generate titles for those abstracts, according to the humor constraints of their paired titles. Two annotators were tasked with rating the higher quality title among the two from different ChatGPT versions, obtaining a Cohen's Kappa score of 0.756 for agreement on 10 common instances.\footnote{If one can not differentiate between the two titles, it is allowed to annotate them as equal.}}

\section{Training on extra parts besides abstract}
\label{app:extra_train}
\ycrev{We do the same filtering 
in \S\ref{sec:data} except for restricting to main conference papers, as there are no venue labels; additionally, we remove the papers which have empty title, abstract, introduction, or conclusion sections in the data.
The filtered data contains 22,452 papers, which are then split into train, dev, and test sets in a ratio of 8:1:1. For ``[MODEL]+X'' models, we concatenate the texts of the three parts by two ``</s>'' tokens as the model input.
For LED models, we limit the maximal input length to 2,048, which is able to cover the concatenated inputs of the great majority of instances; as for BARTXsum, the maximal input length is 1,024, which indicates the inputs of around half of the instances will be truncated.} 

\ycrev{
We train all models using the Trainer API from huggingface with a learning rate of 4e-5 and a batch size of 32 for 20 epochs; the other hyperparameters are default. Each training was stopped by an early stopping with 2 patience, based on the rouge scores on the dev set. We use beam search with 5 beams and a length penalty of 2 for decoding.}

\section{MODEL+A vs.\ MODEL+X}
\label{app:ax}

Table \ref{tab:ax_key} illustrates the examples of abstract-title pairs where the important keywords were missing from the abstracts and only available in other parts like conclusion, and Table \ref{tab:ax_ha} displays the examples of titles with hallucinations.

\begin{table*}[!hbt]
\centering
\small
\begin{tabularx}{\textwidth}{lX}
\toprule
Abstract   & This paper describes a lexicon organized around systematic polysemy: a set of word senses that are related in systematic and predictable ways. The lexicon is derived by a fully automatic extraction method which utilizes a clustering technique called tree-cut. We compare our lexicon to WordNet cousins, and the inter-annotator disagreement observed between WordNet Semcor and DSO corpora.      \citep{tomuro-2001-tree}                                                                                                                                                                                                                                                                                                                                                                                                                                         \\ \midrule
LED+A      & A systematic polysemy lexicon based on tree-cut                                                                                                                                                                                                                                                                                                                                                                                                                                                                                                                                                                                                                                                                                                                                                                                                   \\
LED+X      & A Systematic Polysemy Lexicon Based on Tree-Cut \colorbox{green}{Extraction}                                                                                                                                                                                                                                                                                                                                                                                                                                                                                                                                                                                                                                                                                                                                                                                    \\ \midrule \midrule
Abstract   & We address the problem dealing with a large collection of data, and investigate the use of automatically constructing category hierarchy from a given set of categories to improve classification of large corpora. We use two well-known techniques, partitioning clustering, []-means and a [] to create category hierarchy. []-means is to cluster the given categories in a hierarchy. To select the proper number of [], we use a [] which measures the degree of our disappointment in any differences between the true distribution over inputs and the learner's prediction. Once the optimal number of [] is selected, for each cluster , the procedure is repeated. Our evaluation using the 1996 Reuters corpus which consists of 806,791 documents shows that automatically constructing hierarchy improves classification accuracy. \citep{fukumoto-suzuki-2004-comparison}
\\ \midrule
BARTXsum+A & Automatic Construction of Category Hierarchy for Improved Classification of Large Corpora                                                                                                                                                                                                                                                                                                                                                                                                                                                                                                                                                                                                                                                                                                                                                       \\
BARTXsum+X & Automatic Construction of Category Hierarchy for \colorbox{green}{Text Classification}          \\ \bottomrule
\end{tabularx}
\caption{Examples of abstract-title pairs where the important keywords were missing from the abstracts and only available in other parts like conclusion. We highlight the keywords in the titles from ``[MODEL]+X'' systems. Tokens masked with ``[]'' are those with OCR errors that could not be recognized.}
\label{tab:ax_key}
\end{table*}

\begin{table*}[!hbt]
\centering
\small
\begin{tabularx}{\textwidth}{lX}
\toprule
Paper   & \citet{1178c9db5fcb472299ef40c350f93097}                                                                                                                                                                                                                                                                                                                                                                                                                                         \\ \midrule
LED+A      & Location Disambiguation Using Spatial and Temporal Clues                                                                                                                                                                                                                                                                                                                                                                                                                                                                                                                                                                                                                                                                                                                                                                                       \\
LED+X      & Location Disambiguation Using Spatial \colorbox{pink}{Clustering} and Temporal Consistency                                                                                                                                                                                                                                                                                                                                                                                                                                                                                                                                                                                                                                                                                                                                     \\ \midrule \midrule
Paper   & \citet{10.5555/1859664.1859761}
\\ \midrule
BARTXsum+A & Automatic Disambiguation of Chinese Sentiment Ambiguous Adjectives Using Twitter                                                                                                                                                                                                                                                                                                                                                                                                                                                                                                                                                                                                                                                                                                                                                \\
BARTXsum+X & \colorbox{pink}{NUS-CORE}: Using Twitter to Disambiguate Adjective Sentiment Ambiguous Adjectives          \\ \bottomrule
\end{tabularx}
\caption{Examples of titles with hallucinations. We highlight the hallucinated words in the titles from ``[MODEL]+X'' systems.}
\label{tab:ax_ha}
\end{table*}

\end{document}